\documentclass[lettersize,journal]{IEEEtran}
\usepackage{amsmath,amsfonts}
\usepackage{algorithmic}
\usepackage{algorithm}
\usepackage{array}
\usepackage[caption=false,font=normalsize,labelfont=sf,textfont=sf]{subfig}
\usepackage{textcomp}
\usepackage{stfloats}
\usepackage{url}
\usepackage{booktabs}
\usepackage{verbatim}
\usepackage{graphicx}
\usepackage{subcaption}
\usepackage[colorlinks=true]{hyperref}
\usepackage{cite}
\begin{document}

\title{Efficient VQ-QAT and Mixed Vector/Linear quantized Neural Networks}

\author{Tianrun Gou, Puneet Gupta, \textit{Fellow, IEEE} }

\markboth{}%
{}


\maketitle

\begin{abstract}
In this work, we developed and tested 3 techniques for vector quantization (VQ) based model weight compression. To mitigate codebook collapse and enable end-to-end training, we adopted cosine similarity-based assignment. Building on ideas from attention-based formulations in Differentiable K-Means (DKM), we further improved this approach by using cosine similarity for assignment combined with top-1 sampling and a straight-through estimator, thereby eliminating the need for weighted-average reconstruction. Finally, we investigated the use of differentiable neural architecture search (NAS) to adaptively select layer-wise quantization configurations, further optimizing the compression process. Although our method does not consistently outperform existing approaches across all quantization levels, it provides useful insights into the design trade-offs and behaviors of VQ-based model compression methods.
\end{abstract}


\section{Introduction}
\IEEEPARstart{D}{eep} neural network quantization is a promising solution for reducing model size, memory traffic, and energy consumption, thereby enabling efficient deployment on resource-constrained hardware. Among various quantization methods, vector quantization has been widely explored as a means of compressing an entire model into one or a few shared codebooks, where network weights are represented by compact index bits pointing to codebook entries. Compared to scalar or linear quantization, vector quantization has the protential to enable higher compression ratios and naturally aligns with computation-in-memory (CIM) architectures, as codebook-based weight reuse can significantly reduce memory access and improve data locality. However, this benefit comes at the cost of a more complex quantization and inference pipeline. Also, as the vector length increases, the codebook must represent a higher-dimensional weight space, which often leads to increased quantization error and a noticeable drop in model accuracy. As a result, while vector quantization offers attractive hardware advantages, especially for CIM, its accuracy-efficiency trade-off becomes increasingly challenging for longer vectors.

Quantization-aware training of a vector-quantized model involves the joint optimization of full-precision weights (retained only during training), codebook entries, and the assignments of weights to codebook vectors. In vector quantization, codebook collapse is a severe and well-known challenge. Prior work typically addresses this issue either by introducing additional regularization terms or by employing codebook entry eviction mechanisms. These heuristic designs often require careful hyperparameter tuning and are not directly optimized with respect to the overall task loss.

Compared with existing methods, we summarize our contributions as follows:
\begin{itemize}
    \item We propose a simple, plug-and-play, end-to-end quantization-aware training (QAT) framework for vector quantization of deep neural networks, which eliminates the need for careful temperature tuning as required in Differentiable K-Means (DKM).
    \item For several quantization configurations, our method outperforms state-of-the-art quantization frameworks on ResNet-18 for ImageNet classification, achieving both faster training and better accuracy preservation.
    \item We introduce a unified framework that enables layer-wise selection between linear quantization and vector quantization, allowing more flexible and effective compression strategies.
\end{itemize}

\section{Background and Related Work}
This section introduces the basic concept of model quantization, vector quantization and shortage of existing method. 
\subsection{Quantization}
Quantization compresses neural network parameters by mapping floating‑point weights to a finite set of discrete values, which reduces model size, memory traffic, and even computational cost. Quantization methods can be broadly categorized into uniform and non‑uniform approaches.

In uniform quantization, weight values are projected onto evenly spaced levels. A weight element is transformed into an integer through 

\[
   q = \operatorname{clip}\left( \operatorname{round}\left( \frac{x}{s} \right) + z,0,2^{b} - 1 \right), 
\]

where the scaling factor $s$ and zero-point $z$ are defined as

\[
s = \frac{M - m}{2^{b} - 1}, z = \operatorname{round}\left(-\frac{m}{s}\right)
\]
$M$ and $m$ denote the upper and lower bounds of the quantization range.

In contrast, non‑uniform quantization removes the constraint of evenly spaced levels. The quantized value is chosen from a set of learned or predefined levels that need not be uniformly distributed. Examples include logarithmic quantization, where levels grow exponentially, and codebook‑based quantization, where each weight or weight vector is replaced by an entry in a finite codebook. Formally, a weight w is approximated by $c_i$, where $c_i$ is the codebook entry that minimizes the distance between $w$ and $c_i$

\subsection{Vector Quantization}
\begin{figure}[htbp]
    \centering
    \includegraphics[width=\linewidth]{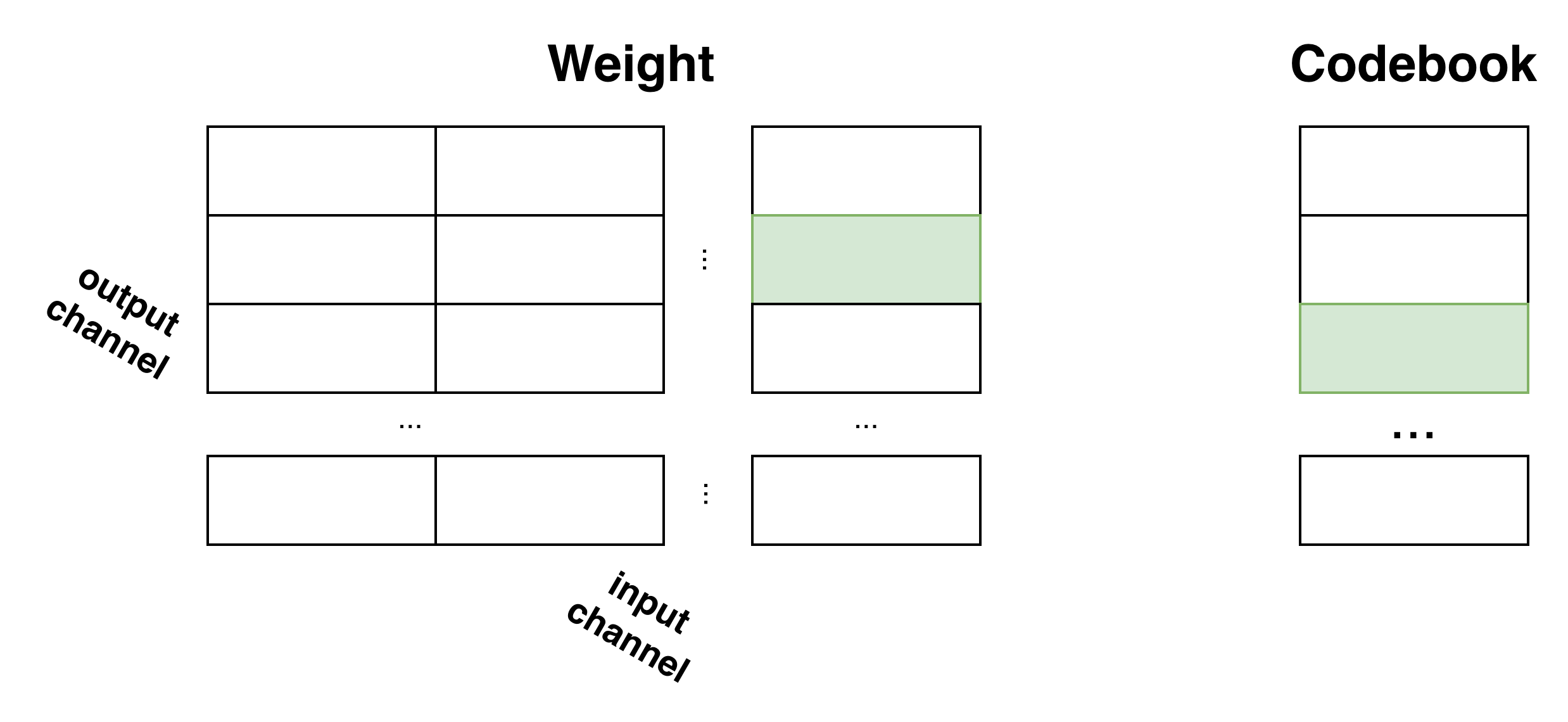}
    \caption{Illustration of codebook based vector quantization for weight}
    \label{fig:vq}
\end{figure}

A special case of codebook quantization is vector quantization, where consecutive elements are grouped into vectors, and mapped to codebook entries which are the vector of the same length, which is shown in Fig \ref{fig:vq}. When the length of vector decreases to 1, it degenerates to scalar codebook quantization. Vector quantization has been studied broudly to compress video, audio and latent space in VAE architecture. 

\subsection{Analog Compute-in-Memory}
In analog compute-in-memory architectures, area and power consumption are typically dominated by the analog-to-digital converters (ADCs), which are placed at the end of each column of the compute array to convert accumulated analog signals into digital values. To mitigate this ADC dominance, one effective approach is to increase the vector length by using taller compute arrays, thereby allowing the compute array to account for a larger fraction of the overall system cost. Moreover, longer vector lengths enable the total dot product to be partitioned into fewer accumulation cycles, reducing the number of ADC conversions required. This not only lowers ADC energy and area overhead but also decreases the cumulative quantization error introduced during repeated analog-to-digital conversions.

\subsection{Ways to mitigate Codebook Collapse}
For vector quantization, a fundamental challenge is codebook collapse, where during training some codebook entries are never assigned to any weight vectors. One approach \cite{ref2} to mitigate this issue is to introduce an additional loss term that explicitly pulls codewords toward the original vectors, which can be interpreted as a form of regularization. Other work \cite{ref3} addresses this problem by evicting inactive codebook entries—i.e., those not assigned to any vectors—and reinitializing them by sampling from the current set of vectors.

\subsection{Differentiable Kmeans and CIMPOOL}
The state-of-the-art approach to vector quantization for weight compression is Differentiable K-Means (DKM)\cite{ref1}, which reformulates traditional k-means clustering as an attention-based soft assignment problem that can be jointly optimized with neural network weights during training. In this formulation, each weight interacts with all codebook entries through a temperature-controlled softmax, resulting in a weighted-average reconstruction of the quantized vector during training. At inference time, however, this soft assignment is collapsed into a hard top-1 assignment by snapping each weight to its nearest centroid. This mismatch between the soft, continuous behavior during training and the hard, discrete behavior at inference introduces a strong dependency on the softmax temperature schedule, requiring careful and often architecture- and configuration-specific tuning. Moreover, while the attention-based formulation allows gradients from a single quantized weight to propagate to all codebook entries, the codebook itself is updated through a heuristic, batch-wise iterative process to ensure convergence during the forward pass, rather than being directly optimized end-to-end under the overall task loss. Together, these factors complicate training and limit the robustness of DKM across different quantization settings.

CIMPool\cite{ref4} mitigates the accuracy degradation that arises from large vector-size compression in compute-in-memory by introducing an auxiliary 1-bit error vector paired with each weight vector, which helps compensate for quantization artifacts due to heavy compression. However, the experimental evaluation in the paper primarily uses the Food-101 classification benchmark, rather than more challenging large-scale vision benchmarks like ImageNet, which limits the strength of the empirical support for the method’s general applicability. Because Food-101 has fewer classes and a smaller dataset size compared to ImageNet-1K, the reported accuracy results do not yet demonstrate how well CIMPool scales to high-complexity, high-variance tasks common in practical vision workloads, leaving room for stronger benchmarking in future work.
\section{Methodology}

This section describes the main contributions of the project. 

\subsection{STE with cosine-similarity-based assignment and projection-based scaling}
\label{subsec:method1}

We first provide the intuition that the directions of the weight vectors are more uniformly distributed than their magnitudes. Consequently, performing codeword assignment  using cosine similarity, which depends only on vector direction, leads to a more balanced assignment of vectors to codebook entries. 

\begin{figure*}[htbp]
    \centering
    \includegraphics[width=0.25\textwidth]{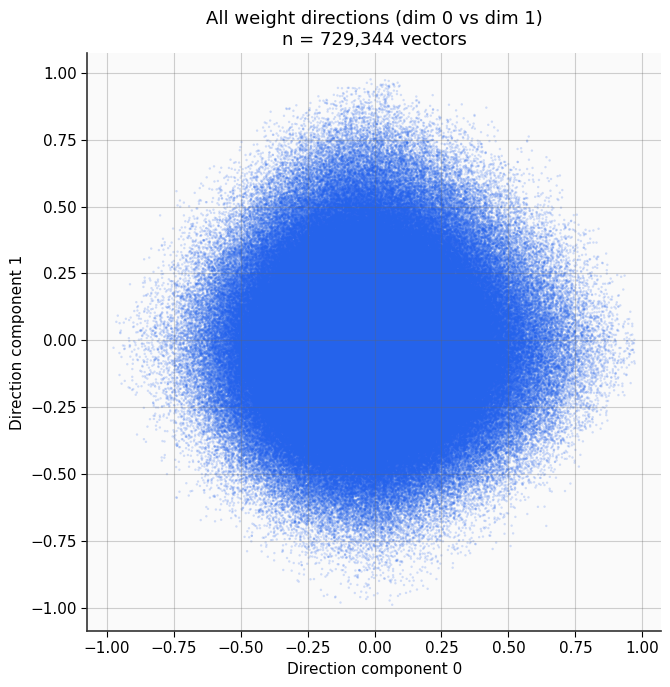}
    \hfill
    \includegraphics[width=0.3\textwidth]{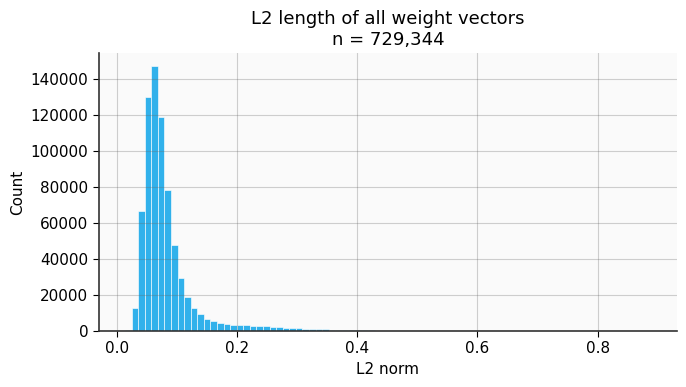}
    \hfill
    \includegraphics[width=0.3\textwidth]{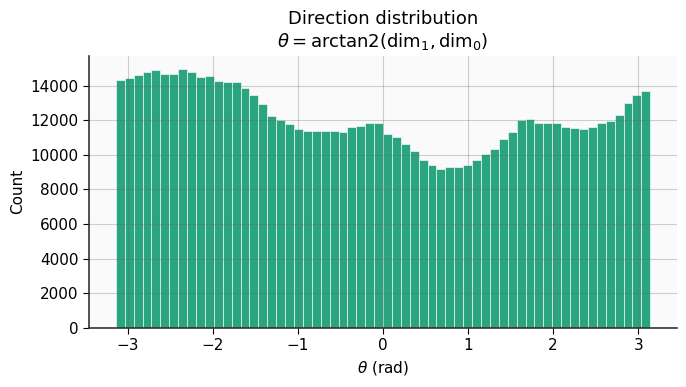}
    \caption{Visualization of weight direction and magnitude distributions.}
    \label{fig:weight_stats}
\end{figure*}

Fig. \ref{fig:weight_stats} visualizes the distribution of pretrained ResNet-18 weight vectors with dimension 2. The left plot shows that weight directions are approximately uniformly distributed, indicating no strong directional bias across the parameter space. This observation is further supported by the nearly flat angular distribution in the right plot. In contrast, the middle plot reveals that the L2 norms of weight vectors are highly imbalanced, with most vectors concentrated at small magnitudes and a long tail of larger values. This decoupling between direction and magnitude suggests that, while directional information is evenly represented, amplitude varies significantly across weights. Motivated by this property, we adopt cosine similarity for codeword assignment to emphasize directional alignment, which helps balance codebook utilization and mitigates codebook collapse in vector quantization.

\begin{figure}[htbp]
    \centering
    \includegraphics[width=\linewidth]{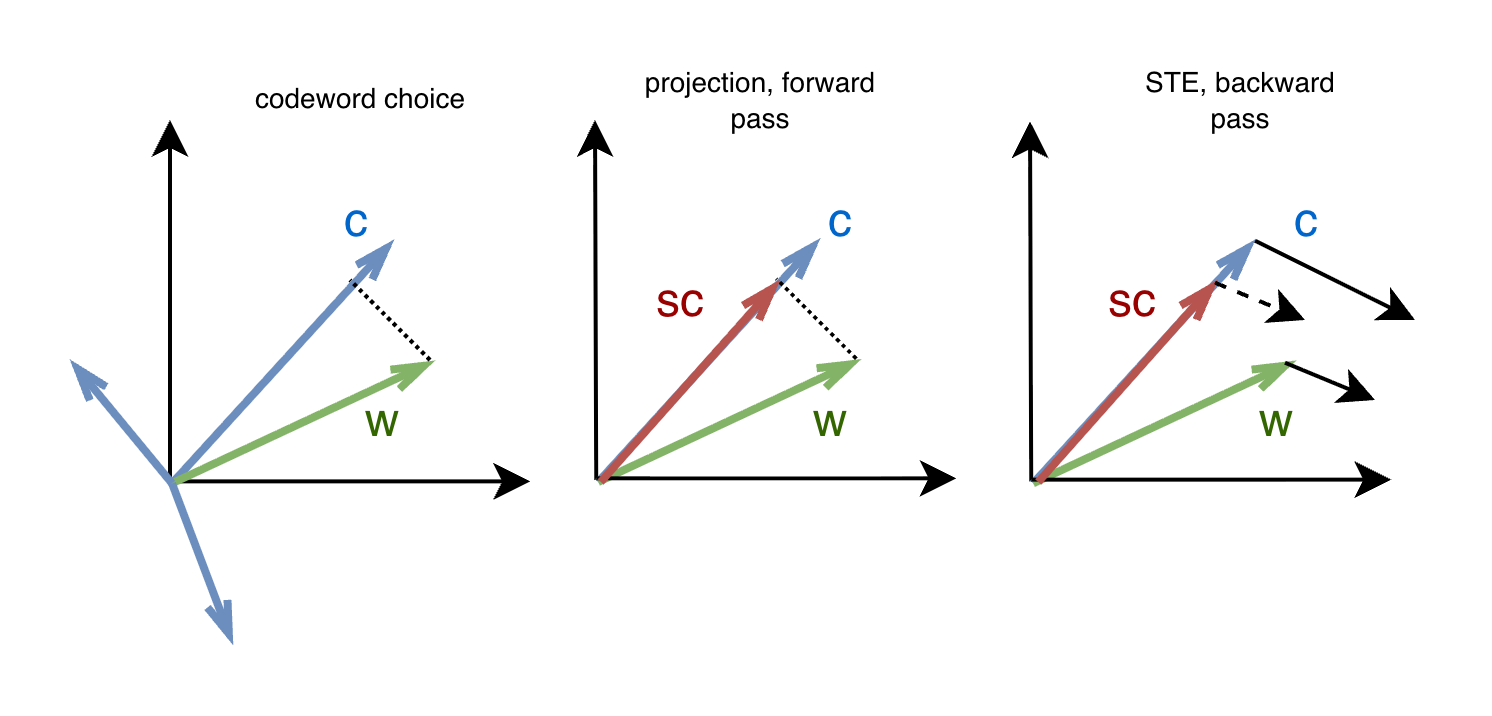}
    \caption{STE with cosine-similarity–based assignment and
projection-based scaling}
    \label{fig:placeholder}
\end{figure}

To preserve magnitude information thus facilitate straight through estimator(STE), we additionally introduce a per-vector scalar to represent the vector length. Specifically, for each weight vector $\mathbf{w} \in \mathbb{R}^d$ and candidate codeword $\mathbf{c} \in \mathcal{C}$, we compute the optimal scalar along the codeword direction:
\[
s = \frac{\mathbf{w}^\top \mathbf{c}}{\|\mathbf{c}\|_2^2}.
\]
The quantized vector here is:
\[
\mathbf{w}_q = s \cdot \mathbf{c},
\]
During the backward pass, a straight-through estimator is applied. Gradients are propagated to $\mathbf{c}$ and $\mathbf{w}$, while the scalar $s$ is treated as a constant:
\[
\frac{\partial \mathcal{L}}{\partial \mathbf{c}} \approx \frac{\partial \mathcal{L}}{\partial \mathbf{w}_q} \cdot s, \quad
\frac{\partial \mathcal{L}}{\partial \mathbf{w}} = \frac{\partial \mathcal{L}}{\partial \mathbf{w_q}}.
\]
This introduces an extra scalar for each weight vector; therefore, the compression ratio with respect to 32-bit floating-point representation is
\[
    CR =\frac{32L}{b_{index}+b_{scalar}}
\]

\subsection{Differentiable Hard Attention based VQ-QAT}
\begin{figure*}[t]
    \centering
    \includegraphics[width=0.95\textwidth]{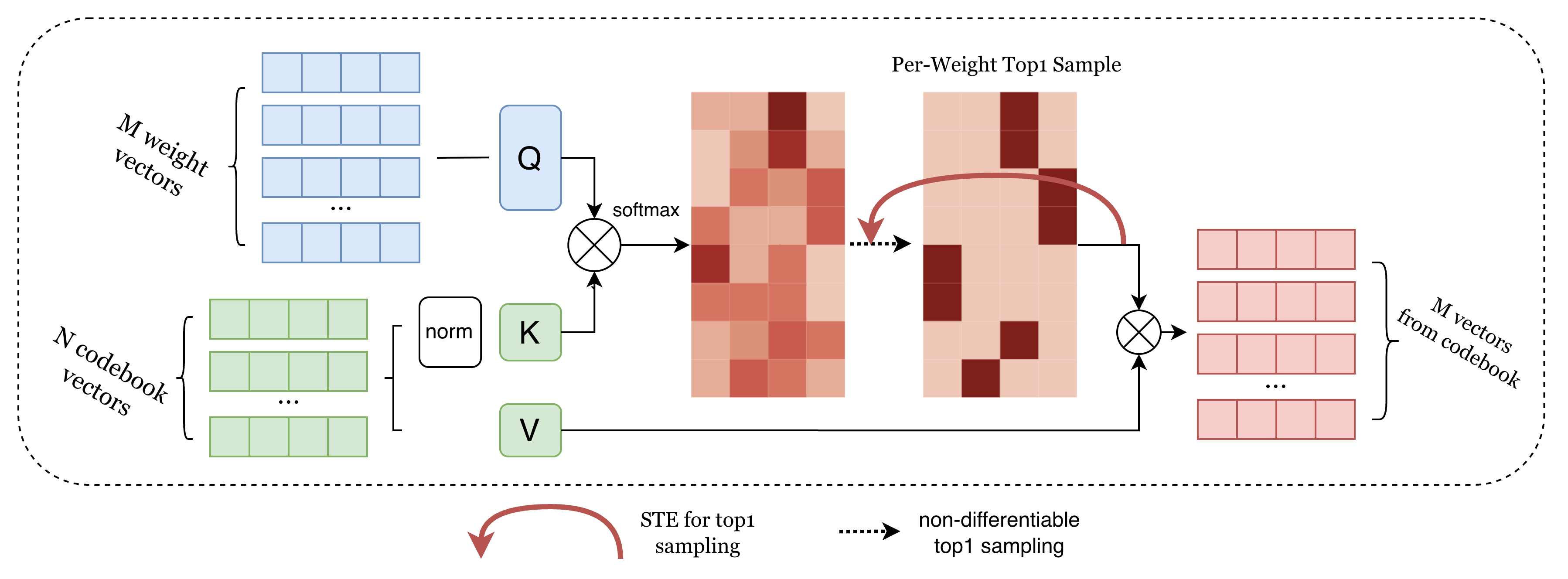}
    \caption{Attention based Vector Quantization aware Training}
    \label{fig:vqqat}
\end{figure*}
Following the idea of cosine-similarity-based assignment, we observe that computing the dot product between each weight vector and the L2-normalized codewords to find similarity and then retrieve codebook entry, naturally resembles an attention operation: the weight vectors act as queries, the normalized codewords serve as keys that encode directional information, and the codeword vectors themselves function as values to be retrieved.
We denote by $w \in \mathbb{R}^d$ a weight vector obtained by grouping consecutive model weights,
and by $\{c_i\}_{i=1}^N$ a learnable codebook. The key vectors are defined as
\begin{equation}
    k_i = \frac{c_i}{\lVert c_i \rVert_2}, \quad i = 1, \dots, N.
\end{equation}
The similarity between $q$ and each codeword is computed as
\begin{equation}
    s_i = q^\top k_i,
\end{equation}
The corresponding assignment probabilities are given by
\begin{equation}
    p_i = \frac{\exp(s_i)}{\sum_{j=1}^L \exp(s_j)}.
\end{equation}
In the forward pass, we use a hard one-hot assignment by selecting the most
probable codeword:
\begin{equation}
    \hat{z}_i =
    \begin{cases}
        1, & i = \arg\max_j p_j, \\
        0, & \text{otherwise},
    \end{cases}
\end{equation}
and obtain the quantized vector
\begin{equation}
    \hat{w} = \sum_{i=1}^L \hat{z}_i \, c_i.
\end{equation}
Due to the use of one-hot sampling, the probability distribution does not participate in the forward pass, which blocks gradient flow and prevents learning. To overcome this, we employ a straight-through estimator, treating the sampled vector $\hat{z}$ as if it were the original probability distribution $p$ during the backward pass. Concretely, gradients are propagated directly from the one-hot vector to the underlying probabilities, enabling end-to-end optimization. This method ensures that training and inference are consistent, removing the need for careful temperature tuning in the softmax.
\\
With a straight-through estimator (STE), the one-hot vector $z$ is treated as
$z \approx p$ during backpropagation. Hence,
\[
\frac{\partial z_i}{\partial p_j} \approx \delta_{ij},
\qquad
\frac{\partial \mathcal{L}}{\partial p_i}
\approx
\frac{\partial \mathcal{L}}{\partial z_i}.
\]
Since $\hat{w} = z^\top V$, we obtain
\[
\frac{\partial \mathcal{L}}{\partial p_i}
\approx
\left(\frac{\partial \mathcal{L}}{\partial \hat{w}}\right)^\top v_i.
\]
Thus, the gradient received by each probability 
$p_i$ is exactly the dot
product between the upstream gradient and codeword $v_i$.
Intuitively, each
codeword is encouraged or discouraged depending on how well it aligns with the
desired weight update. Even though the forward pass selects only one codeword,
every probability $p_i$ receives a meaningful gradient signal, enabling
end-to-end optimization.

Geometrically, the gradient $g$ indicates the direction in weight space that most
reduces the loss. Changing $p_i$ perturbs the quantized vector $\hat{w}$ toward
the direction of $v_i$. If $v_i$ is orthogonal to $g$, this perturbation lies in a
direction that does not affect the loss to first order. Consequently, the
probability $p_i$ receives no gradient signal, and only codewords whose directions
have a non-zero projection onto $g$ (i.e., $g^\top v_i \neq 0$) participate in the
update.

By combining cosine-similarity attention with top-1 sampling and STE, we bypass the need for weighted averages in attention, ensuring exactly consistent behavior between training and inference, while retaining full differentiability.

\subsection{ProxylessNas based layer-wise VQ/LQ selection}
Aside from quantizing the entire network using VQ, another approach is to assign different quantization configurations to different parts of the model. A simple starting point is to apply this on a layer-wise basis.

\begin{figure}[htbp]
    \centering
    \includegraphics[width=1\linewidth]{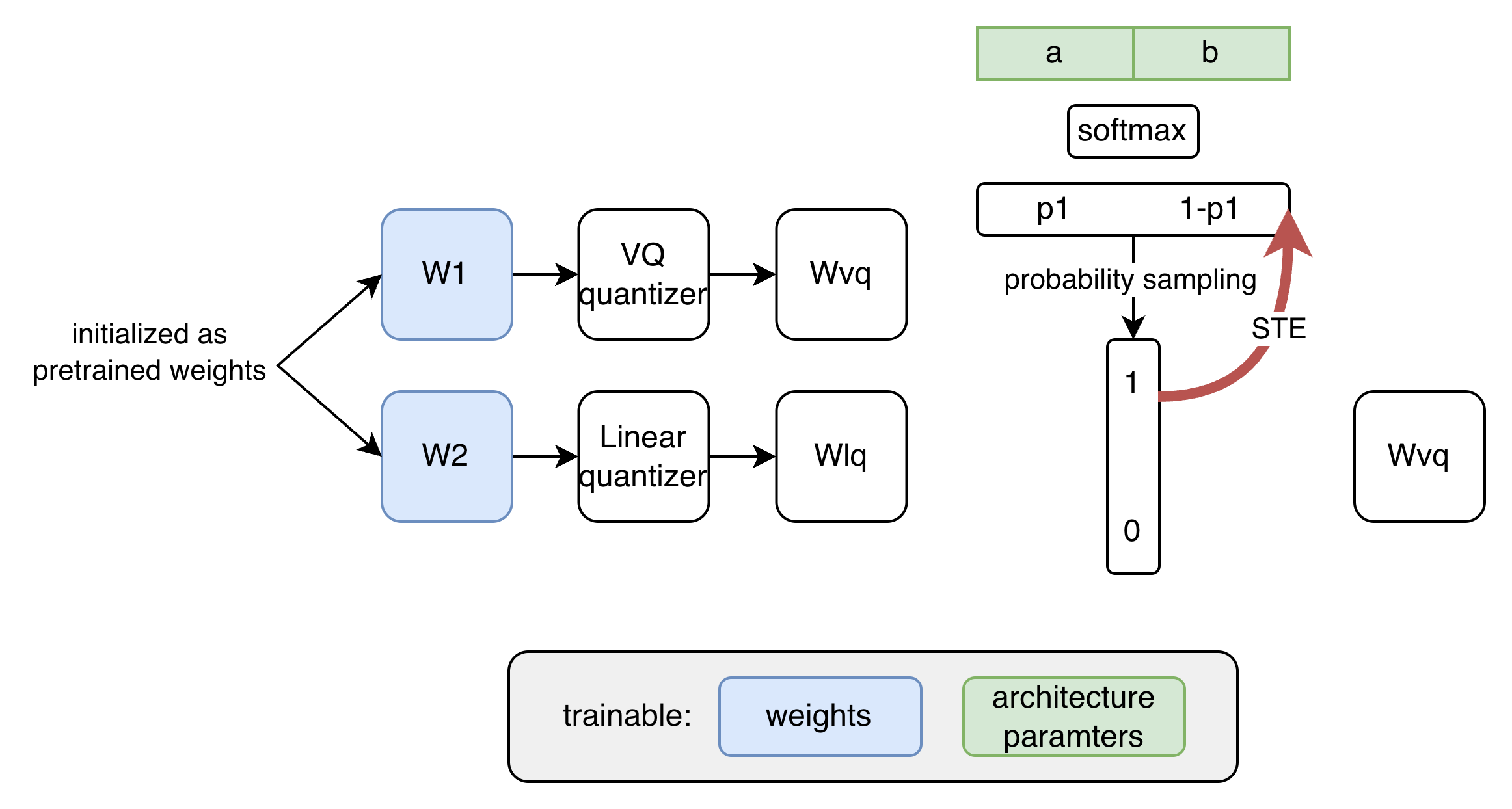}
    \caption{Illustration of ProxylessNas for VQ/LQ selection}
    \label{fig:nas}
\end{figure}

Fig.\ref{fig:nas} illustrates a learnable mixed-quantization framework in which the model dynamically chooses between two different weight quantization strategies during training. Rather than fixing the quantization method a priori, the architecture allows the selection process itself to be optimized jointly with the model parameters, enabling adaptive trade-offs between accuracy and efficiency.

Two parallel weight branches are maintained in the model. In the first branch, the weights $(W_1)$ are quantized by  vector quantization, producing quantized weights denoted as ($W_{\mathrm{vq}}$). In the second branch, the weights ($W_2$) are quantized by a linear quantizer, producing quantized weights ($W_{\mathrm{lq}}$). These two branches represent alternative quantization schemes that can be selected during training.

The selection between the two quantized branches is governed by a set of learnable architecture parameters, labeled (a) and (b). These parameters are passed through a softmax function to produce a probability distribution ($[p_1,, 1 - p_1]$), where $(p_1)$ corresponds to the probability of choosing the VQ-quantized branch. A probabilistic sampling step then draws a binary decision: selecting the VQ branch when the sampled value is 1, or the linear-quantized branch when the sampled value is 0. Since the binary sampling operation is non-differentiable, a straight-through estimator (STE) is employed to enable end-to-end training. During the forward pass, a hard binary decision is applied to select one quantization branch. During the backward pass, gradients are propagated through the sampling operation to the softmax probabilities as if the operation were continuous, allowing meaningful gradient updates to the architecture parameters. This follows the method in \cite{ref5}.

To take parameters into consideration, an extra term is added to the total loss,

\[
L = L_{CE}+\lambda ||w||^2_2 + \beta \mathbb{E}[storage]
\]
where $\mathbb{E}[storage]$ for a certain layer is defined as expectation of total bit storage 
\[
\mathbb{E}[storage] = p\frac{N}{L}Q_{vq} + (1-p)Q_{lq}N
\]

Overall, the framework jointly optimizes two types of trainable components. The model weights associated with each quantization branch are trainable, as indicated by the blue blocks in the figure. In addition, the architecture parameters that control the quantization choice are also trainable, as indicated by the green blocks. This design allows the model to learn both effective weight representations and an appropriate quantization strategy in a unified, differentiable manner.

\section{Experiments}
\subsection{Network architectures}
To evaluate the performance of our method, we conducted quantization-aware training (QAT) on ResNet-18\cite{ref9} for the ImageNet-1K classification task. In all experiments, the first 7x7 convolutional layer was kept in floating-point precision. For the ProxylessNAS component, we also maintained the final linear layer in floating-point precision. All other layers were fully quantized.
\subsection{Initialization}
We loaded pretrained model from \cite{ref6}. The codebook in initialized by performing Kmeans on weight vectors.
\subsection{Training hyper-parameter}
For all experiments, we employed automatic mixed precision (AMP) in PyTorch and trained the models on a single NVIDIA RTX 4090 GPU with a batch size of 128. Network parameters were optimized using stochastic gradient descent (SGD) with a momentum of 0.9. A weight decay of 1e-4 was applied for regularization to mitigate overfitting. The learning rate was individually tuned for each experiment to ensure optimal convergence. A cosine annealing learning rate scheduler was used, decaying the learning rate to zero by the end of training. All models were trained for 100 epochs. For scalar linear quantization and the linear weight quantization used in the ProxylessNAS component, we adopt the quantizer described in~\cite{ref8}, in which the clipping threshold is learned during training.

\section{Results}
\subsection{STE with cosine-similarity-based assignment and projection-based scaling}
We compare our results with DKM and EWGS \cite{ref7}, one of the existing state-of-the-art linear QAT work.
\begin{table}[htbp]
\centering
\caption{Top-1 accuracy (\%) on ImageNet-1K for EWGS, quantization methods.}
\label{tab:imagenet_ewgs}
\setlength{\tabcolsep}{8pt} 
\begin{tabular}{lccc}
\toprule
\textbf{Bit-width} & \textbf{DKM } & \textbf{EWGS (w/o linear layer)} & \textbf{Ours} \\
\midrule
8/(4+4) (1bit) & 67.8\%  & 67.0\% & 63.9\% \\
16/(8+8) (1bit)   & 67.8\%  & 67.0\%  & 63.9\%\\
32/(8+8) (0.5bit)   & 65.5\%  & N/A\%  & 56.3\% \\

\bottomrule
\end{tabular}
\end{table}
Table~\ref{tab:imagenet_ewgs} shows that introducing an additional scalar does not close the performance gap to DKM or EWGS. For a fixed vector length (e.g., 8), partitioning the bit budget between the index and a scalar parameter reduces the number of representable codewords compared to allocating all bits to the index alone, as the former constitutes a strict subset of the latter. These results indicate that, under the evaluated bit budgets, preserving directional resolution is more advantageous than dedicating bits to magnitude.

\subsection{Differentiable Hard Attention based VQ-QAT}
For this part, the scalar are removed and all bits are for index. The training flow follows Fig.\ref{fig:vqqat}
\begin{table}[htbp]
\centering
\caption{Top-1 accuracy (\%) on ImageNet-1K for various quantization methods. 
}\label{tab:imagenet_methods}
\setlength{\tabcolsep}{8pt} 
\begin{tabular}{lccc}
\toprule
\textbf{Bit-width} & \textbf{DKM } & \textbf{EWGS (w/o linear layer)} & \textbf{Ours} \\
\midrule
4/8 (2bit)   & 68.9\%  & 69.6\%  & \textbf{68.66\%} \\
4/4 (1bit)   & 67.0\%  & 67.0\%  & 63.80\% \\
8/8 (1bit)   & 67.8\%  & --      & \textbf{67.08\%} \\
16/8 (1/2bit) & 65.5\% & N/A      & 62.60\% \\
\bottomrule
\end{tabular}
\end{table}

As shown in Table~\ref{tab:imagenet_methods}, our hard-attention VQ-QAT matches the baselines at moderate bit widths and remains competitive as the budget decreases. At 4/8 and 8/8, our method is within 0.3--0.8\% of the best reported accuracy, while avoiding the soft-to-hard mismatch in DKM. When the bit budget is reduced to 4/4 or 16/8, all methods degrade, but the relative ordering is preserved, indicating that the proposed assignment remains stable under tighter compression.

The training of our method is significantly more efficient than that of DKM. While DKM relies on a per-batch iterative K-means procedure, which results in approximately 40 minutes per epoch on the hardware we carry experiments on, our approach completes each epoch in only 6 minutes by eliminating the need for iterative clustering. Furthermore, DKM requires careful tuning of the temperature for each dataset and quantization configuration, which substantially increases both the total training time and the associated hardware costs. In contrast, our method achieves comparable or better performance with markedly reduced computational overhead.

\subsection{ProxylessNas based layer-wise VQ/LQ selection}
In this stage, we impose a global bitwidth budget for the network. Once the average bitwidth falls below the specified budget, the architecture is fixed, and only the selected branches are further trained.
\begin{table}[htbp]
\centering
\caption{Top-1 accuracy (\%) on ImageNet-1K for VQ/LQ NAS}
\label{tab:imagenet_vqlq_nas}
\setlength{\tabcolsep}{8pt} 
\begin{tabular}{lcccc}
\toprule
\textbf{VQ} &\textbf{LQ}& \textbf{Final Average Bitwidth}  & \textbf{Accuracy} \\
\midrule
16/8 (0.5bit) & 4 & 0.85 & 63.89\%   \\
64/8 (0.125bit) & 8 & 1.34 & 61.91\%\\
64/8 (0.125bit) & 4 & 1.34& 63.74\%  \\

\bottomrule
\end{tabular}
\end{table}

\begin{figure*}[htbp]
    \centering
    \includegraphics[width=0.8\textwidth]{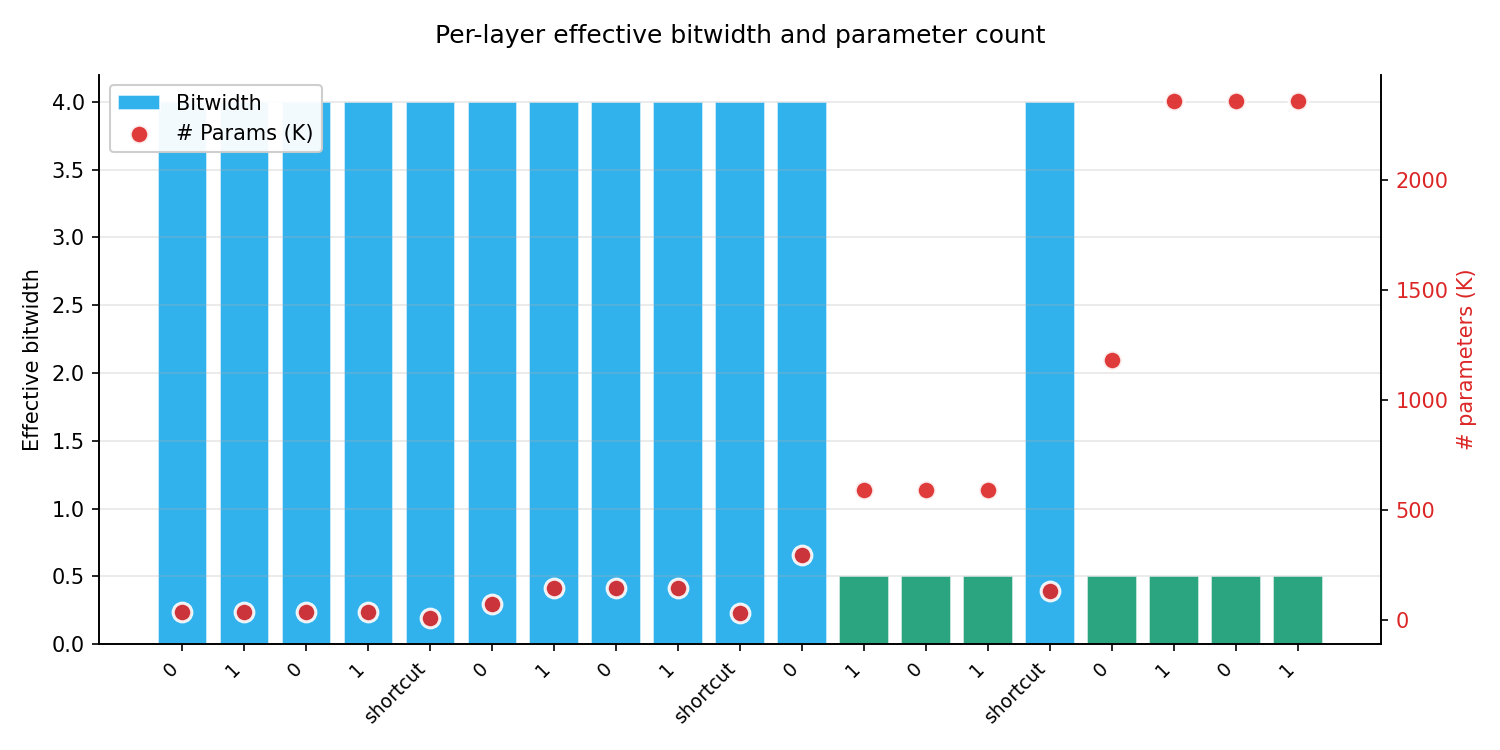}
    \caption{Search result for a given budget of 1bit, between VQ of 8bit index for vector length of 16 and 4bit linear quanization}
    \label{fig:nas_result}
\end{figure*}

As shown in Fig.~\ref{fig:nas_result}, layers closer to the output are assigned lower bitwidths, and vector quantization is applied to improve the overall compression ratio. Since ProxylessNAS is employed, the quantization configuration is optimized end-to-end, jointly considering both compression and model performance. However, as shown in \ref{tab:imagenet_vqlq_nas}, under the same ultra-low bitwidth, this approach does not outperform vanilla linear quantization or our formally proposed method. Additionally, we observe that vector quantization with shorter vector lengths, but a smaller total average bitwidth, can yield better performance. We attribute this to the excessive information loss that occurs when using long vector quantization; layers quantized with long vectors can degrade overall model performance significantly.

\section{Conclusion}
In this project, we explored 3 mechnisms for vector quantization based QAT for model weight compression. We first developed scalar envolved STE, also hard attention based flow, and found that it's better to assign all bits to the direction itself as we go into longer vector quantization case. However, we found it's rather hard to maintain accuracy of the model as vector length goes to 32 or longer.

Owing to the use of one-hot sampling combined with a straight-through estimator (STE), our method eliminates the need for weighted-average reconstruction during training, resulting in identical computational graphs for training and inference. This alignment removes the necessity for temperature-based hyperparameter tuning, as the training and inference behaviors are mathematically consistent through the STE. Moreover, the codebook is directly optimized via gradient descent on the task loss, without requiring iterative clustering procedures. Consequently, our framework significantly reduces the computational overhead per training step.

In the ProxylessNAS component, joint optimization of architecture parameters and network weights enables the model to automatically protect sensitive layers by assigning them linear quantization, while aggressively compressing less critical layers using vector quantization (VQ). We further observe that employing excessively long VQ vectors can substantially degrade overall performance, even though doing so allows more layers to remain at higher linear quantization precision under the same bitwidth budget.


\section*{Acknowledgments}
We thank George Karfakis for valuable discussions and advice on this project.

\end{document}